\documentclass[runningheads]{llncs}
\usepackage{graphicx}
\usepackage{xcolor}
\usepackage{amsmath}
\usepackage{amssymb}
\usepackage{multirow} 
\usepackage{bigstrut}
\usepackage{tabularx}
\usepackage{wrapfig}
\usepackage{float}
\usepackage{marvosym}
\begin{document}

\title{Boosting MLPs with a Coarsening Strategy for Long-Term Time Series Forecasting}
\titlerunning{Boosting MLPs with a Coarsening Strategy}
%
\author{Nannan Bian \inst{1} \and
Minhong Zhu \inst{2} \and
Li Chen \inst{3} \and
Weiran Cai \inst{1} \textsuperscript{(\Letter)}}
%
\authorrunning{N. Bian et al.}
%

\institute{School of Computer Science and Technology, Soochow University, \\
Suzhou 215006, China \\ \email{wrcai@suda.edu.cn} \and
School of Biology \& Basic Medical Science, Soochow University, \\
Suzhou 215006, China \and
School of Physics and Information Technology, Shaanxi Normal University,\\
Xi’an 710061, China}
%
\maketitle              
\begin{abstract}

Deep learning methods have been exerting their strengths in long-term time series forecasting. However, they often struggle to strike a balance between expressive power and computational efficiency. Resorting to multi-layer perceptrons (MLPs) provides a compromising solution, yet they suffer from two critical problems caused by the intrinsic point-wise mapping mode, in terms of deficient contextual dependencies and inadequate information bottleneck. Here, we propose the Coarsened Perceptron Network (CP-Net), featured by a coarsening strategy that alleviates the above problems associated with the prototype MLPs by forming information granules in place of solitary temporal points. The CP-Net utilizes primarily a two-stage framework for extracting semantic and contextual patterns, which preserves correlations over larger timespans and filters out volatile noises. This is further enhanced by a multi-scale setting, where patterns of diverse granularities are fused towards a comprehensive prediction. Based purely on convolutions of structural simplicity, CP-Net is able to maintain a linear computational complexity and low runtime, while demonstrates an improvement of 4.1\% compared with the SOTA method on seven forecasting benchmarks. Code is available at https://github.com/nannanbian/CPNet  

\end{abstract}
\section{Introduction}
Long-term multivariate time series forecasting, encompassing the prediction of future changes over an extended period using a substantial amount of historical data, finds diverse applications in the real world. Examples include weather forecasting, traffic flow prediction, economic planning, and electricity demand forecasting. Given the non-linear modeling capabilities of neural networks, recent studies concentrate on capturing intricate time patterns by employing deep learning methods  \cite{zhou2021informer,wu2021autoformer} as substitutes for traditional statistical approaches \cite{ariyo2014stock,taylor2018forecasting} in real world time series analyses. 

Similar to that in the language processing domain, time series are composed of temporal patterns, where each time point may not only form short-term dependencies with adjacent time points, such as hourly relations in the consumption of electricity, but also jointly constitute long-term global dependencies with other distant ones, such as the quarterly or yearly variations in the above case. However, in the short-term aspect, the unique nature of time series, being a collection of continuously recorded values arranged in the temporal order, implies that a single time point often lacks adequate information for analysis. Yet in the long-term aspect, when dealing with long input sequences, time series forecasting shares the same efficiency challenge as language modelling. Hence, an efficient deep learning model is consistently in pursuit that is able to comprehend both short- and long-term temporal patterns in a compound way, while maintaining a low computational complexity.

Facing the dilemma, simple frameworks resorting to linear- or MLP-based layers provide a potential solution for capturing long-term dependencies while preserving a linear computational complexity. It traces back to DLinear that questions the necessity of attention mechanism and achieves comparable performances by simply adopting a linear layer in association with timescale separation \cite{zeng2023transformers}. This makes the use of linear- or MLP-based layers a new trend that substitutes Transformers in time series forecasting. Yet, by operating point-wise mapping modes, these architectures generally suffer from two critical problems, i.e., they are deficient in preserving contextual dependencies and inadequate to form an information bottleneck for filtering out redundant noises. Therefore, engaging an effective enhancing strategy will be necessary for such frameworks.  

To solve the aforementioned problems, we propose \textbf{C}oarsened \textbf{P}erceptron \textbf{Net}work (CP-Net), a two-stage framework composed of multi-scale Token Projection Blocks and Contextual Sampling Blocks. The main motivation is to employ a coarsening scheme that forms granularity of information to address the limitations of point-wise mapping of the MLP layer. This is to ensure rich semantic and contextual patterns to be detected in the time series and unwanted volatile information is adequately removed.

Centered on the functionality of the MLP, the enhancing scheme introduces enhancement both prior and posterior to the point-wise projection. It has two key advantages: On one hand, both stages are designed in the spirit of preserving crucial temporal correlations. The Token Projection Block transfers the input signal to tokens that encompass indispensable semantic information which is missing in the point-wise MLP projection; whereas the Contextual Sampling Block, featuring a combination of dilated and equi-convolutions, forms a down-sampling functionality that supplements contextual patterns in the output of the MLP projection. On the other hand, the information contained in granularity aids to filter out noises that would otherwise be present in the point-wise mapping. To further preserve the temporal patterns of diverse granularity \cite{wang2022micn,shabani2022scaleformer}, we implement this boosting strategy in a multi-scale setting by employing different token lengths and sampling rates, which are merged towards a comprehensive final prediction. Notably, all components sandwiching the core MLP in this architecture are purely convolution-based, which renders the advantage of an overall linear computational complexity.

The main contributions of our work are summarized as follows:
\begin{itemize}
    \item[$\bullet$] We propose CP-Net featuring a two-stage coarsening strategy that alleviates the intrinsic drawbacks of the point-wise mapping of MLPs and boosts their efficiency in prediction tasks. The backbone architecture consists of sequentially arranged Token Projection Blocks and Contextual Sampling Blocks to extract semantic and contextual temporal patterns, respectively, which provides further an adequate information bottleneck for noise filtering.
    \item[$\bullet$] Our enhancing components are purely convolution-based, which are paralleled in multiple branches to integrate the temporal patterns of different granularity. This architectural simplicity renders the computation within a linear complexity and low runtime.
    \item[$\bullet$] Experiments on seven popular multivariate long-term forecasting benchmarks justify the proposed strategy with improvements over the state-of-the-art method, which achieves a decrease of 4.1\% and 3.3\% on MSE and MAE (averaged over four prediction lengths), respectively.
\end{itemize}

\section{Related Work}

\subsection{Convolution-based Forecasting Approaches}

The framework of convolutional neural networks (CNNs) features a filtering kernel that is crafted to capture local features optimally in the temporal signals. While employing CNNs holds the potential to enhance the precise capture of local patterns, challenges may arise in grasping global ones. TCN \cite{bai2018empirical} is the first attempt in this direction which employs causal convolution to avoid leakage of future information and extends the receptive field of convolution kernels with dilated convolution to model longer-range dependencies. SCINet \cite{liu2022scinet} goes for a diverse path by utilizing convolution filters to extract features from the downsampled subsequences. MICN \cite{wang2022micn} further improves the idea of downsampling, which leverages downsampling convolution and isometric convolution to extract local features and global correlation; it further introduces a multi-scale structure to capture patterns in different time scales. More recently, TimesNet \cite{wu2022timesnet} introduces a novel 2D convolution architecture to capture intraperiod- and interperiod-variations simultaneously. By converting 1D time series into a collection of 2D tensors based on multiple periods, it effectively mitigates the limitation of previous downsampling methods acting on 1D time series. However, despite considerable efforts in optimizing long-range patterns capturing, CNN-based methods are still limited by its receptive field of the kernels leading to incomplete awareness of distant time points. Regarding both the proficiency in extracting local information and low complexity, this work takes full advantages of convolutional layers throughout the enhancing strategy.

\subsection{Linear- and MLP-based Forecasting Approaches}
The recent work DLinear \cite{zeng2023transformers} relies solely on simple linear model and surpasses most Transformer methods. DLinear employs only one linear layer for each component after decomposing the time series into trend and seasonal components. This motivates further exploration into utilizing linear layers for time series forecasting to reduce computational complexity while preserving effectiveness. Prior to this, there have been many studies using the MLP structure for time series forecasting. N-HiTS \cite{challu2023nhits} is an extension of the famous N-Beats \cite{oreshkin2019n} model for long-term time series forecasting, which solves the problem of the volatility of the predictions and their computational complexity by incorporating novel hierarchical interpolation and multi-rate data sampling techniques. LightTS \cite{zhang2022less} applies an MLP-based structure to implement interval sampling and continuous sampling strategies to reflect the dependencies of the original sequence. FreTS \cite{yi2024frequency} transforms sequences into the frequency domain, redesigning the MLP to separately compute the real and imaginary parts of the frequency components. The superiority in the structural simplicity and the long-term scope encourages us to also utilize the MLP architecture as the core projection functionality. Yet, our work differs from existing models by adopting a coarsening strategy to solve the intrinsic shortcoming in the point-wise mode of mapping of MLPs, based on the idea of forming information granules.

\section{Methods}

For multivariate time series forecasting with $N$ different variables, given a historical sequence $\mathbf{X} \in \mathbb{R}^{I\times N} $ of $I$ steps, our goal is to predict the future $O$ time steps $\mathbf{Y} \in \mathbb{R}^{O\times N}$. We target at a scalable approach for the effective comprehension of both long- and short-term temporal patterns. In this section, we introduce the CP-Net, a multi-scale framework that features a purely convolution-based coarsening strategy, composed of Token Projection Blocks and Contextual Sampling Blocks. 
These components are crucial for restoring important short-term patterns to the global point-wise projection of the MLP layer, ensuring a comprehensive understanding of the time series.

\begin{figure*}[t]
    \hspace{-0.6cm}
    \includegraphics[width=1.1\textwidth]{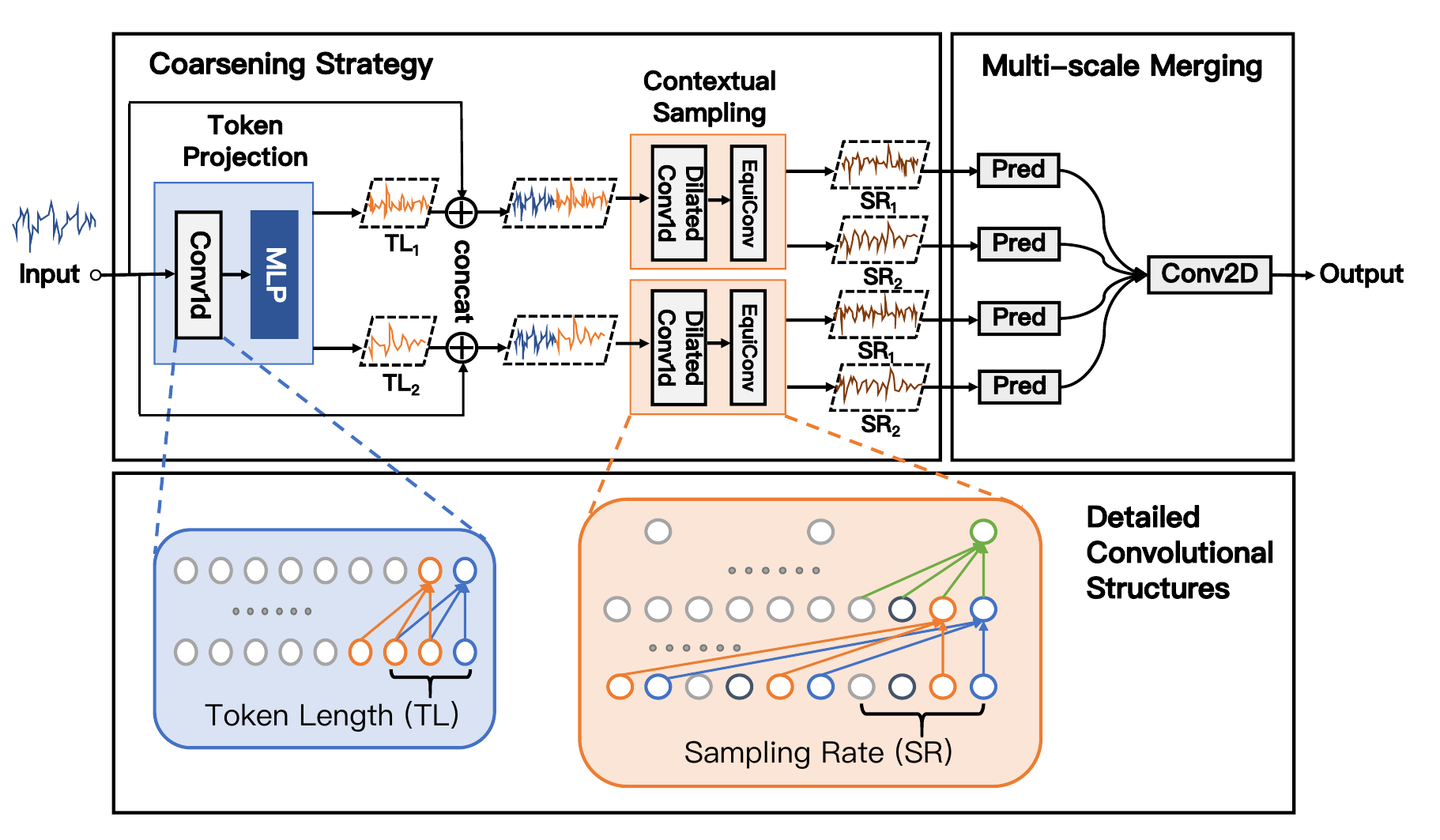}
    \caption{Overview of CP-Net. Two-stage coarsening strategy: Time points in input signals are coarsened prior to the projection of the MLP layer with a Token Projection Block as to render a preliminary prediction. Posterior to that, short-term correlations are further extracted with a Contextual Sampling Block. Multi-scale merging: The multi-branch setting decodes and fuses the output information of diverse granularities to render a compound prediction. Detailed convolutional structures: The token Projection Block aggregates semantic information by employing a standard convolution, whereas the Contextual Sampling Block incorporates temporal dependencies and filters out volatile noises by proper down-sampling through dilated and equispaced convolutions.}
    \label{fig:framework}
\end{figure*}

\subsection{Overall Structure}

The two-stage coarsening framework employs a Token Projection Block and a Contextual Sampling Block sequentially to preserve important temporal correlations at the input and output of the MLP layer, respectively, as illustrated in Fig. \ref{fig:framework}. The Token Projection Block realizes a global projection to construct a preliminary prediction, for which the input signals are aggregated to form coarsened tokens prior to the MLP layer. Here, the token length ($TL$), representing the number of time points used to aggregate coarse-grained tokens, is used to determine the input granularity. Subsequently, the Contextual Sampling Block establishes correlations over a time span that rejoins the historical input with the preliminary output of the MLP. It also provides a down-sampling function to preserve significant patterns while mitigate noises. The granularity of information here depends on the sampling rate ($SR$).

Accounting for temporal patterns of diverse granularities in real-world scenarios, we employ a multi-scale strategy to integrate them. We parallelize different branches of Token Projection Blocks and Contextual Sampling Blocks with diverse predefined pairs of $TL$ and $SR$. Specifically, for a branch with $TL_i$ and $SR_j$, given the time series $\mathbf{X_{in}}$ normalized by Instance Normalization \cite{ulyanov2016instance,kim2021reversible} as input, the temporal patterns through the bi-stage coarsening process can be computed as
\begin{equation}\label{equ:branch}
  \begin{split}  {\mathbf{X_m^{i,j}}}=\operatorname{CS^{j}}\left(\operatorname{TP^{i}}\left(\mathbf{X_{in}}\right)\right), \\
  \end{split}
\end{equation}
\noindent where $\operatorname{TP^{i}(\cdot)}$ and $\operatorname{CS^{j}(\cdot)}$ denotes the Token Projection Block with $TL=TL_i$ and Contextual Sampling Block with $SR=SR_j$, respectively.

Various temporal patterns are ultimately fused to achieve a final prediction through a multi-scale merging approach. Specifically, separate predictors are used to decode the temporal patterns of different branches to achieve the desired prediction length, and the final prediction is obtained through a merging function. Each branch utilizes a pair of $TL$ and $SR$ to detect correlations at different granularity levels. The merging process can be represented as
\begin{equation}\label{equ:merge}
  \begin{split}
  {\mathbf{Y}}=\operatorname{Merge}\left(\mathbf{X_m^{i,j}}\right), \\
  \end{split}
\end{equation}
where $\operatorname{Merge}(\cdot)$ is the multi-scale merging approach mentioned above.
It is worth noting that we utilize the channel-independent assumption similar to that in \cite{nie2022time}. In the following subsections, we detail the designs of comprising blocks.

\subsection{Token Projection Block}
The Token Projection Block provides granular information as the input to the MLP. It takes the normalized raw time series as input and outputs a preliminary future prediction through the MLP mapping of constructed tokens. As shown in Fig. \ref{fig:framework}, the process begins with transforming the input sequence into overlapping consecutive coarse-grained tokens, which is to behold semantic information in the local patterns. An MLP projects these coarse-grained tokens into preliminary future predictions, providing a preliminary insight into the future trends of the time series data.
  
Concretely, let $\mathbf{X_{in}} \in \mathbb{R}^{I\times N}$ be the input series and $TL$ the length of the coarse-grained tokens, the preliminary prediction $\mathbf{X_{tp}} \in \mathbb{R}^{O\times N}$ can be computed as
\begin{equation}\label{equ:Conv1dandMLP}
  \begin{split}
  {\mathbf{X_{tp}}}= 
  {\operatorname{TP(\mathbf{X_{in}})}}= 
  \operatorname{MLP}\left(\operatorname{Conv1d}\left(\mathbf{X_{in}}\right) \right),  \\
  \end{split}
\end{equation}
where $\operatorname{Conv1d}(\cdot)$ denotes a standard 1D convolution layer with its kernel size equal to $TL$, and $\operatorname{MLP}(\cdot)$ is a 2-layer perceptron realizing the global projection.

\subsection{Contextual Sampling Block}

The Contextual Sampling Block forms information granules from the MLP output. Consecutive time points are aggregated and down-sampled, which generates new coarse-grained representations that retain essential temporal information while effectively filter out redundant noises. 
As shown in Fig. \ref{fig:framework}, we first use the dilated convolution to capture contextual temporal patterns in a periodic manner, with the dilation rate is the sampling rate $SR$, which is to better capture longer-range dependencies. In order to avoid information loss caused by zero-padding and to maintain closer connection with the historical sequence, we reuse the latest time steps from the input historical sequence as padding elements and concatenate them with the preliminary prediction from the previous block. This approach allows us to extract temporal patterns while preserve the historical information. The above process can be formulated as
\begin{equation}\label{equ:dalitedconvolution}
  \begin{split}  {\mathbf{{X}_{cs}}}=\operatorname{DilatedConv1d}\left(\operatorname{Concat}(\mathbf{X_{tp}, \mathbf{X_{in}}})\right) \\
  \end{split}
\end{equation}
\noindent where $\mathbf{{X}_{cs}} \in \mathbb{R}^{(I+O) \times N}$, $\operatorname{DilatedConv1d}(\cdot)$ is a 1D dilated convolution layer, and $\operatorname{Concat}(\cdot)$ is the specialized padding strategy reusing the historical series.   
  
Subsequently, equispaced convolution, a type of convolution where the kernel size is equal to the stride, is employed for down-sampling contextual temporal patterns. The down-sampling does not only help preserve crucial longer-range temporal relationships from the previous step but further filters out redundant information, addressing the issue of information bottleneck related to prototype MLPs. Again with the sampling rate $SR$, the final contextual temporal pattern can be expressed as
\begin{equation}\label{equ:equiconvolution}
  \begin{split}
  {\mathbf{X_{m}}}=\operatorname{EquiConv1d}\left(\mathbf{{X}_{cs}}\right) \\
  \end{split}
\end{equation}

\noindent where $\operatorname{EquiConv1d}(\cdot)$ is the equispaced convolution with a kernel size $SR$ and $\mathbf{X_{m}} \in \mathbb{R}^{M\times N}$, with $M = (I+O) / SR$ representing the length of the output representation. In all, by linking far-spanned temporal points and down-sampling, the two convolution layers contribute to carving out clear key patterns and function as an adequate information bottleneck.

\subsection{Multi-Scale Merging}

With a multi-scale setting, we integrate temporal patterns at all granularity levels. We parallelize the Token Projection Blocks and Contextual Sampling Blocks in multiple branches corresponding to diverse token lengths and sampling rates. The outputs are fused through a convolution-based merging approach. We define $\mathbf{S}=\{(TL_i,SR_j)\}$ as the ensemble of parameters for all such branches containing a Token Projection Block and a Contextual Sampling Block (values are set separately for simplicity). Multiple branches corresponding to the pairs defined in $\mathbf{S}$ can be computed in parallel, generating temporal patterns at different scales. Since information granules are preserved in different branches, they are decoded independently. Specifically, separate predictors are employed to decode the temporal patterns in each branch
\begin{equation}\label{equ:Predictor}
  \begin{split}
  {\mathbf{Y_{m}^{i,j}}}= 
  \operatorname{Predictor}\left(\mathbf{X_{m}^{i,j}}\right),  \\
  \end{split}
\end{equation}

\noindent where $\operatorname{Predictor}(\cdot)$ is a 2-layer perceptron. $\mathbf{X_{m}^{i,j}} \in \mathbb{R}^{M\times N}, M = (I+O) / SR_j$ is the output of a given branch of the Contextual Sampling Block. $\mathbf{Y_{m}^{i,j}} \in \mathbb{R}^{(I+O)\times N}$ represents a corresponding future prediction. We utilize 2D convolution to blend the branch-wise predictions $\mathbf{Y_{m}^{i,j}}$ with proper weights, which are truncated to the desired output length $O$ and compared with the ground truth in a supervised training. With the given parameters $\mathbf{S}$, the final prediction is computed as 
\begin{equation}\label{equ:Conv2d}
  \begin{split}
  {\mathbf{Y}}=\operatorname{Truncate}\left(\underset{(i, j) \in \mathbf{S}}{\operatorname{Conv2d}}\left(\mathbf{Y_{m}^{i,j}}\right)\right) \\
  \end{split}
\end{equation}
\noindent where $ \mathbf{Y} \in \mathbb{R}^{O \times N}$ represents the resulted final prediction.

\section{Experiments}
\subsection{Multivariate Long-term Time Series Forecasting}
\subsubsection{Datasets}
We evaluate the performance of our proposed CP-Net on seven datasets, including ETT \cite{zhou2021informer} (ETTm1, ETTm2, ETTh1, ETTh2), Electricity, Traffic and Weather. These datasets have been widely used as benchmarks, whose public splits and evaluation standards are available on \cite{wu2021autoformer}. The statistics of those datasets are shown in Table \ref{tab:dataset}.
\begin{table}[H]
  \centering
  \caption{Statistics of seven commonly used benchmark datasets. }
    \begin{tabular}{c|c|c|c|c|c|c|c}
    \hline
    dataset & ETTm1 & ETTm2 & ETTh1 & ETTh2 & Electricity & Traffic & Weather \bigstrut\\
    \hline
    variates & 7     & 7     & 7     & 7     & 321   & 862   & 21  \bigstrut\\
    \hline
    time steps & 69680 & 69680 & 17420 & 17420 & 26304 & 17544 & 52696 \bigstrut\\
    \hline
    granularity & 15 mins & 15 mins & Hourly & Hourly & Hourly & Hourly & 10 mins  \bigstrut\\
    \hline
    \end{tabular}%
  \label{tab:dataset}%
\end{table}%
\subsubsection{Baseline Models and Setup}
We choose representative models from three categories as our baselines, including the CNN-based model TimesNet \cite{wu2022timesnet}, the Transformer-based models PatchTST \cite{nie2022time}, FEDformer \cite{zhou2022fedformer}, Autoformer \cite{wu2021autoformer}, and the Linear- and MLP-based models DLinear \cite{zeng2023transformers} and LightTS \cite{zhang2022less}.

All of the models follow the same experimental setups with the same look-back window $I=96$ and four prediction lengths $O \in \{96,192,336,720\}$. We collect the baseline results from TimesNet \cite{wu2022timesnet} with a look-back window $I=96$. For PatchTST, we run the officially provided code with default hyper-parameter settings and a different look-back window $I=96$ from the original paper and reported the results. We employ the commonly used MSE and MAE as the metrics for evaluation.

\subsubsection{Main Results}

For multivariate long-term time series forecasting, our model outperforms or on par with the baselines on all benchmarks, as displayed in Table \ref{tab:forecasting_results}. In comparison with the current SOTA model Transformer-based PatchTST, our model with the simple architecture demonstrates superior performance with a 4.1\% reduction in MSE and a 3.3\% reduction in MAE. On large datasets such as Traffic and Electricity, the proposed model consistently outperforms PatchTST in all settings. Compared with the previous best CNN-based model TimesNet, our model achieves a 6.6\% reduction in MSE and a 3.3\% reduction in MAE. Notably on the largest Traffic dataset, our model's MSE is 18.2\% lower. In contrast to the top-performing linear model DLinear, our model surpasses it significantly with a 14.4\% decrease in MSE and an 11.6\% decrease in MAE. This clearly demonstrates that our coarsening strategy has effectively alleviated the drawbacks of point-wise projections and thus significantly outperforms the model in the same category.

\begin{table}[H]
  \centering
  \begin{small}
  \caption{Multivariate long-term time series forecasting results with CP-Net and baseline models. We set the input length $I=96$, and forecasting length $O \in \{96,192,336,720\}$. The lower the MSE or MAE, the better the results, with the best results highlighted in \textbf{bold} and the second best \underline{underlined}.}\label{tab:forecasting_results}
   \begin{tabular}{c|c|cc|cc|cc|cc|cc|cc|cc}
    \hline
    \multicolumn{2}{c|}{Models} & \multicolumn{2}{c|}{\scalebox{0.84}{Ours}} & \multicolumn{2}{c|}{\scalebox{0.84}{TimesNet}} & \multicolumn{2}{c|}{\scalebox{0.84}{PatchTST}} & \multicolumn{2}{c|}{\scalebox{0.84}{Dlinear}} & \multicolumn{2}{c|}{\scalebox{0.84}{LightTS}} & \multicolumn{2}{c|}{\scalebox{0.84}{FEDformer}} & \multicolumn{2}{c}{\scalebox{0.84}{Autoformer}} \bigstrut\\
    \hline
    \multicolumn{2}{c|}{Metric} & \scalebox{0.84}{MSE}   & \scalebox{0.84}{MAE}   & \scalebox{0.84}{MSE}   & \scalebox{0.84}{MAE}   & \scalebox{0.84}{MSE}   & \scalebox{0.84}{MAE}   & \scalebox{0.84}{MSE}   & \scalebox{0.84}{MAE}   & \scalebox{0.84}{MSE}   & \scalebox{0.84}{MAE}   & \scalebox{0.84}{MSE}   & \scalebox{0.84}{MAE}   & \scalebox{0.84}{MSE}   & \scalebox{0.84}{MAE} \bigstrut\\
    \hline
    \multirow{4}[2]{*}{\rotatebox{90}{ETTm1}} & 96    & \scalebox{0.84}{\textbf{0.321}} & \scalebox{0.84}{\textbf{0.359}} & \scalebox{0.84}{\underline{0.338}}  & \scalebox{0.84}{0.375}  & \scalebox{0.84}{0.339}  & \scalebox{0.84}{\underline{0.368}}  & \scalebox{0.84}{0.345}  & \scalebox{0.84}{0.372}  & \scalebox{0.84}{0.374}  & \scalebox{0.84}{0.400}  & \scalebox{0.84}{0.379}  & \scalebox{0.84}{0.419}  & \scalebox{0.84}{0.505}  & \scalebox{0.84}{0.475} \bigstrut[t]\\
          & 192   & \scalebox{0.84}{\textbf{0.367}} & \scalebox{0.84}{\textbf{0.383}} & \scalebox{0.84}{\underline{0.374}}  & \scalebox{0.84}{\underline{0.387}}  & \scalebox{0.84}{\underline{0.374}}  & \scalebox{0.84}{0.388}  & \scalebox{0.84}{0.380}  & \scalebox{0.84}{0.389}  & \scalebox{0.84}{0.400}  & \scalebox{0.84}{0.407}  & \scalebox{0.84}{0.426}  & \scalebox{0.84}{0.441}  & \scalebox{0.84}{0.553}  & \scalebox{0.84}{0.496}  \\
          & 336   & \scalebox{0.84}{\textbf{0.400}} & \scalebox{0.84}{\underline{0.406}} & \scalebox{0.84}{0.410}  & \scalebox{0.84}{0.411}  & \scalebox{0.84}{\underline{0.406}}  & \scalebox{0.84}{\textbf{0.405}} & \scalebox{0.84}{0.413}  & \scalebox{0.84}{0.413}  & \scalebox{0.84}{0.438}  & \scalebox{0.84}{0.438}  & \scalebox{0.84}{0.445}  & \scalebox{0.84}{0.459}  & \scalebox{0.84}{0.621}  & \scalebox{0.84}{0.537}  \\
          & 720   & \scalebox{0.84}{\textbf{0.462}} & \scalebox{0.84}{\underline{0.441}}  & \scalebox{0.84}{0.478}  & \scalebox{0.84}{0.450}  & \scalebox{0.84}{\textbf{0.462}}  & \scalebox{0.84}{\textbf{0.440}} & \scalebox{0.84}{\underline{0.474}}  & \scalebox{0.84}{0.453}  & \scalebox{0.84}{0.527}  & \scalebox{0.84}{0.502}  & \scalebox{0.84}{0.543}  & \scalebox{0.84}{0.490}  & \scalebox{0.84}{0.671}  & \scalebox{0.84}{0.561}  \bigstrut[b]\\
    \hline
    \multirow{4}[2]{*}{\rotatebox{90}{ETTm2}} & 96    & \scalebox{0.84}{\textbf{0.176}} & \scalebox{0.84}{\textbf{0.259}} & \scalebox{0.84}{\underline{0.187}}  & \scalebox{0.84}{0.267}  & \scalebox{0.84}{\textbf{0.176}} & \scalebox{0.84}{\underline{0.260}}  & \scalebox{0.84}{0.193}  & \scalebox{0.84}{0.292}  & \scalebox{0.84}{0.209}  & \scalebox{0.84}{0.308}  & \scalebox{0.84}{0.203}  & \scalebox{0.84}{0.287}  & \scalebox{0.84}{0.255}  & \scalebox{0.84}{0.339}  \bigstrut[t]\\
          & 192   & \scalebox{0.84}{\textbf{0.241}} & \scalebox{0.84}{\textbf{0.300}} & \scalebox{0.84}{0.249}  & \scalebox{0.84}{0.309}  & \scalebox{0.84}{\underline{0.242}} & \scalebox{0.84}{\underline{0.303}}  & \scalebox{0.84}{0.284}  & \scalebox{0.84}{0.362}  & \scalebox{0.84}{0.311}  & \scalebox{0.84}{0.382}  & \scalebox{0.84}{0.269}  & \scalebox{0.84}{0.328}  & \scalebox{0.84}{0.281}  & \scalebox{0.84}{0.340}  \\
          & 336   & \scalebox{0.84}{\textbf{0.299}} & \scalebox{0.84}{\textbf{0.337}} & \scalebox{0.84}{0.321}  & \scalebox{0.84}{0.351}  & \scalebox{0.84}{\underline{0.302}}  & \scalebox{0.84}{\underline{0.342}}  & \scalebox{0.84}{0.369}  & \scalebox{0.84}{0.427}  & \scalebox{0.84}{0.442}  & \scalebox{0.84}{0.466}  & \scalebox{0.84}{0.325}  & \scalebox{0.84}{0.366}  & \scalebox{0.84}{0.339}  & \scalebox{0.84}{0.372}  \\
          & 720   & \scalebox{0.84}{\textbf{0.397}} & \scalebox{0.84}{\textbf{0.394}} & \scalebox{0.84}{0.408}  & \scalebox{0.84}{0.403}  & \scalebox{0.84}{\underline{0.399}}  & \scalebox{0.84}{\underline{0.396}}  & \scalebox{0.84}{0.554}  & \scalebox{0.84}{0.522}  & \scalebox{0.84}{0.675}  & \scalebox{0.84}{0.587}  & \scalebox{0.84}{0.421}  & \scalebox{0.84}{0.415}  & \scalebox{0.84}{0.433}  & \scalebox{0.84}{0.432}  \bigstrut[b]\\
    \hline
    \multirow{4}[2]{*}{\rotatebox{90}{ETTh1}} & 96    & \scalebox{0.84}{\underline{0.380}} & \scalebox{0.84}{\textbf{0.397}} & \scalebox{0.84}{0.384}  & \scalebox{0.84}{0.402}  & \scalebox{0.84}{0.410}  & \scalebox{0.84}{0.416}  & \scalebox{0.84}{0.386}  & \scalebox{0.84}{\underline{0.400}}  & \scalebox{0.84}{0.424}  & \scalebox{0.84}{0.432}  & \scalebox{0.84}{\textbf{0.376}}  & \scalebox{0.84}{0.419}  & \scalebox{0.84}{0.449}  & \scalebox{0.84}{0.459}  \bigstrut[t]\\
          & 192   & \scalebox{0.84}{\underline{0.435}}  & \scalebox{0.84}{\textbf{0.429}} & \scalebox{0.84}{0.436}  & \scalebox{0.84}{\textbf{0.429}} & \scalebox{0.84}{0.459}  & \scalebox{0.84}{0.444}  & \scalebox{0.84}{0.437}  & \scalebox{0.84}{\underline{0.432}}  & \scalebox{0.84}{0.475}  & \scalebox{0.84}{0.462}  & \scalebox{0.84}{\textbf{0.420}} & \scalebox{0.84}{0.448}  & \scalebox{0.84}{0.500}  & \scalebox{0.84}{0.482}  \\
          & 336   & \scalebox{0.84}{\underline{0.479}}  & \scalebox{0.84}{\textbf{0.450}} & \scalebox{0.84}{0.491}  & \scalebox{0.84}{0.469}  & \scalebox{0.84}{0.500}  & \scalebox{0.84}{0.465}  & \scalebox{0.84}{0.481}  & \scalebox{0.84}{\underline{0.459}}  & \scalebox{0.84}{0.518}  & \scalebox{0.84}{0.488}  & \scalebox{0.84}{\textbf{0.459}} & \scalebox{0.84}{0.465}  & \scalebox{0.84}{0.521}  & \scalebox{0.84}{0.496}  \\
          & 720   & \scalebox{0.84}{\textbf{0.490}} & \scalebox{0.84}{\textbf{0.473}} & \scalebox{0.84}{0.521}  & \scalebox{0.84}{0.500}  & \scalebox{0.84}{\underline{0.498}}  & \scalebox{0.84}{\underline{0.487}}  & \scalebox{0.84}{0.519}  & \scalebox{0.84}{0.516}  & \scalebox{0.84}{0.547}  & \scalebox{0.84}{0.533}  & \scalebox{0.84}{0.506}  & \scalebox{0.84}{0.507}  & \scalebox{0.84}{0.514}  & \scalebox{0.84}{0.512}  \bigstrut[b]\\
    \hline
    \multirow{4}[2]{*}{\rotatebox{90}{ETTh2}} & 96    & \scalebox{0.84}{\textbf{0.291}} & \scalebox{0.84}{\textbf{0.343}} & \scalebox{0.84}{0.340}  & \scalebox{0.84}{0.374}  & \scalebox{0.84}{\underline{0.302}}  & \scalebox{0.84}{\underline{0.348}}  & \scalebox{0.84}{0.333}  & \scalebox{0.84}{0.387}  & \scalebox{0.84}{0.397}  & \scalebox{0.84}{0.437}  & \scalebox{0.84}{0.358}  & \scalebox{0.84}{0.397}  & \scalebox{0.84}{0.346}  & \scalebox{0.84}{0.388}  \bigstrut[t]\\
          & 192   & \scalebox{0.84}{\textbf{0.367}} & \scalebox{0.84}{\textbf{0.392}} & \scalebox{0.84}{0.402}  & \scalebox{0.84}{0.414}  & \scalebox{0.84}{\underline{0.388}}  & \scalebox{0.84}{\underline{0.400}}  & \scalebox{0.84}{0.477}  & \scalebox{0.84}{0.476}  & \scalebox{0.84}{0.520}  & \scalebox{0.84}{0.504}  & \scalebox{0.84}{0.429}  & \scalebox{0.84}{0.439}  & \scalebox{0.84}{0.456}  & \scalebox{0.84}{0.452}  \\
          & 336   & \scalebox{0.84}{\textbf{0.412}} & \scalebox{0.84}{\textbf{0.426}} & \scalebox{0.84}{0.452}  & \scalebox{0.84}{0.452}  & \scalebox{0.84}{\underline{0.421}}  & \scalebox{0.84}{\underline{0.431}}  & \scalebox{0.84}{0.594}  & \scalebox{0.84}{0.541}  & \scalebox{0.84}{0.626}  & \scalebox{0.84}{0.559}  & \scalebox{0.84}{0.496}  & \scalebox{0.84}{0.487}  & \scalebox{0.84}{0.482}  & \scalebox{0.84}{0.486}  \\
          & 720   & \scalebox{0.84}{\textbf{0.420}} & \scalebox{0.84}{\textbf{0.440}} & \scalebox{0.84}{0.462}  & \scalebox{0.84}{0.468}  & \scalebox{0.84}{\underline{0.429}}  & \scalebox{0.84}{\underline{0.445}}  & \scalebox{0.84}{0.831}  & \scalebox{0.84}{0.657}  & \scalebox{0.84}{0.863}  & \scalebox{0.84}{0.672}  & \scalebox{0.84}{0.463}  & \scalebox{0.84}{0.474}  & \scalebox{0.84}{0.515}  & \scalebox{0.84}{0.511}  \bigstrut[b]\\
    \hline
    \multirow{4}[2]{*}{\rotatebox{90}{Electricity}} & 96    & \scalebox{0.84}{\underline{0.180}}  & \scalebox{0.84}{\textbf{0.264}} & \scalebox{0.84}{\textbf{0.168}} & \scalebox{0.84}{\underline{0.272}}  & \scalebox{0.84}{0.196}  & \scalebox{0.84}{0.285}  & \scalebox{0.84}{0.197}  & \scalebox{0.84}{0.282}  & \scalebox{0.84}{0.207}  & \scalebox{0.84}{0.307}  & \scalebox{0.84}{0.193}  & \scalebox{0.84}{0.308}  & \scalebox{0.84}{0.201}  & \scalebox{0.84}{0.317}  \bigstrut[t]\\
          & 192   & \scalebox{0.84}{\underline{0.186}} & \scalebox{0.84}{\textbf{0.269}} & \scalebox{0.84}{\textbf{0.184}} & \scalebox{0.84}{0.289}  & \scalebox{0.84}{0.198}  & \scalebox{0.84}{0.289}  & \scalebox{0.84}{0.196}  & \scalebox{0.84}{\underline{0.285}}  & \scalebox{0.84}{0.213}  & \scalebox{0.84}{0.316}  & \scalebox{0.84}{0.201}  & \scalebox{0.84}{0.315}  & \scalebox{0.84}{0.222}  & \scalebox{0.84}{0.334}  \\
          & 336   & \scalebox{0.84}{\underline{0.202}}  & \scalebox{0.84}{\textbf{0.286}} & \scalebox{0.84}{\textbf{0.198}} & \scalebox{0.84}{\underline{0.300}}  & \scalebox{0.84}{0.214}  & \scalebox{0.84}{0.304}  & \scalebox{0.84}{0.209}  & \scalebox{0.84}{0.301}  & \scalebox{0.84}{0.230}  & \scalebox{0.84}{0.333}  & \scalebox{0.84}{0.214}  & \scalebox{0.84}{0.329}  & \scalebox{0.84}{0.231}  & \scalebox{0.84}{0.338}  \\
          & 720   & \scalebox{0.84}{\underline{0.243}}  & \scalebox{0.84}{\textbf{0.319}} & \scalebox{0.84}{\textbf{0.220}} & \scalebox{0.84}{\underline{0.320}}  & \scalebox{0.84}{0.256}  & \scalebox{0.84}{0.336}  & \scalebox{0.84}{0.245}  & \scalebox{0.84}{0.333}  & \scalebox{0.84}{0.265}  & \scalebox{0.84}{0.360}  & \scalebox{0.84}{0.246}  & \scalebox{0.84}{0.355}  & \scalebox{0.84}{0.254}  & \scalebox{0.84}{0.361}  \bigstrut[b]\\
    \hline
    \multirow{4}[2]{*}{\rotatebox{90}{Traffic}} & 96    & \scalebox{0.84}{\textbf{0.492}} & \scalebox{0.84}{\underline{0.325}} & \scalebox{0.84}{0.593}  & \scalebox{0.84}{\textbf{0.321}}  & \scalebox{0.84}{\underline{0.557}}  & \scalebox{0.84}{0.365}  & \scalebox{0.84}{0.650}  & \scalebox{0.84}{0.396}  & \scalebox{0.84}{0.615}  & \scalebox{0.84}{0.391}  & \scalebox{0.84}{0.587}  & \scalebox{0.84}{0.366}  & \scalebox{0.84}{0.613}  & \scalebox{0.84}{0.388}  \bigstrut[t]\\
          & 192   & \scalebox{0.84}{\textbf{0.492}} & \scalebox{0.84}{\textbf{0.322}} & \scalebox{0.84}{0.617}  & \scalebox{0.84}{\underline{0.336}}  & \scalebox{0.84}{\underline{0.545}}  & \scalebox{0.84}{0.356}  & \scalebox{0.84}{0.598}  & \scalebox{0.84}{0.370}  & \scalebox{0.84}{0.601}  & \scalebox{0.84}{0.382}  & \scalebox{0.84}{0.604}  & \scalebox{0.84}{0.373}  & \scalebox{0.84}{0.616}  & \scalebox{0.84}{0.382}  \\
          & 336   & \scalebox{0.84}{\textbf{0.506}} & \scalebox{0.84}{\textbf{0.328}} & \scalebox{0.84}{0.629}  & \scalebox{0.84}{\underline{0.336}}  & \scalebox{0.84}{\underline{0.555}}  & \scalebox{0.84}{0.359}  & \scalebox{0.84}{0.605}  & \scalebox{0.84}{0.373}  & \scalebox{0.84}{0.613}  & \scalebox{0.84}{0.386}  & \scalebox{0.84}{0.621}  & \scalebox{0.84}{0.383}  & \scalebox{0.84}{0.622}  & \scalebox{0.84}{0.337}  \\
          & 720   & \scalebox{0.84}{\textbf{0.539}} & \scalebox{0.84}{\textbf{0.345}} & \scalebox{0.84}{0.640}  & \scalebox{0.84}{\underline{0.350}}  & \scalebox{0.84}{\underline{0.592}}  & \scalebox{0.84}{0.376}  & \scalebox{0.84}{0.645}  & \scalebox{0.84}{0.394}  & \scalebox{0.84}{0.658}  & \scalebox{0.84}{0.407}  & \scalebox{0.84}{0.626}  & \scalebox{0.84}{0.382}  & \scalebox{0.84}{0.660}  & \scalebox{0.84}{0.408}  \bigstrut[b]\\
    \hline
    \multirow{4}[2]{*}{\rotatebox{90}{Weather}} & 96    & \scalebox{0.84}{\underline{0.177}}  & \scalebox{0.84}{\textbf{0.217}} & \scalebox{0.84}{\textbf{0.172}} & \scalebox{0.84}{0.220}  & \scalebox{0.84}{0.179}  & \scalebox{0.84}{\underline{0.219}}  & \scalebox{0.84}{0.196}  & \scalebox{0.84}{0.255}  & \scalebox{0.84}{0.182}  & \scalebox{0.84}{0.242}  & \scalebox{0.84}{0.217}  & \scalebox{0.84}{0.296}  & \scalebox{0.84}{0.266}  & \scalebox{0.84}{0.336}  \bigstrut[t]\\
          & 192   & \scalebox{0.84}{\underline{0.226}}  & \scalebox{0.84}{\textbf{0.259}} & \scalebox{0.84}{\textbf{0.219}} & \scalebox{0.84}{\underline{0.261}}  & \scalebox{0.84}{\underline{0.226}}  & \scalebox{0.84}{\textbf{0.259}}  & \scalebox{0.84}{0.237}  & \scalebox{0.84}{0.296}  & \scalebox{0.84}{0.227}  & \scalebox{0.84}{0.287}  & \scalebox{0.84}{0.276}  & \scalebox{0.84}{0.336}  & \scalebox{0.84}{0.307}  & \scalebox{0.84}{0.367}  \\
          & 336   & \scalebox{0.84}{\underline{0.281}}  & \scalebox{0.84}{\textbf{0.298}} & \scalebox{0.84}{\textbf{0.280}} & \scalebox{0.84}{\underline{0.306}}  & \scalebox{0.84}{\textbf{0.280}} & \scalebox{0.84}{\textbf{0.298}} & \scalebox{0.84}{0.283}  & \scalebox{0.84}{0.335}  & \scalebox{0.84}{0.282}  & \scalebox{0.84}{0.334}  & \scalebox{0.84}{0.339}  & \scalebox{0.84}{0.380}  & \scalebox{0.84}{0.359}  & \scalebox{0.84}{0.395}  \\
          & 720   & \scalebox{0.84}{0.357}  & \scalebox{0.84}{\textbf{0.348}} & \scalebox{0.84}{0.365}  & \scalebox{0.84}{\underline{0.359}}  & \scalebox{0.84}{0.355}  & \scalebox{0.84}{\textbf{0.348}}  & \scalebox{0.84}{\textbf{0.345}} & \scalebox{0.84}{0.381}  & \scalebox{0.84}{\underline{0.352}}  & \scalebox{0.84}{0.386}  & \scalebox{0.84}{0.403}  & \scalebox{0.84}{0.428}  & \scalebox{0.84}{0.419}  & \scalebox{0.84}{0.428}  \bigstrut[b]\\
    \hline
    \end{tabular}%
    \end{small}
\end{table}%

\subsection{Ablation Study}
To verify the effectiveness of our proposed coarsening strategy in the model backbone, we conduct an ablation study on different variants of the model. Considering that the main purpose is to verify the coarsening module, we retain the MLP layer and independently remove the semantic coarsening and contextual coarsening modules independently. For clarity, we refer to TP and CS as the two coarsening modules. Concretely, we test the performance of the following four variants:
\begin{itemize}
    \item[$\bullet$] \textbf{CP-Net}: represents the standard model we propose.
    \item[$\bullet$] \textbf{w/o TP}: represents removing coarsening in the TP block. 
    \item[$\bullet$] \textbf{w/o CS}: represents removing coarsening in the CS block.
    \item[$\bullet$] \textbf{w/o TP\&CS}: represents that both coarsening modules are removed only with the MLP left.
\end{itemize}

\begin{table}[H]
   \centering
   \setlength{\tabcolsep}{7pt} 
   \caption{Ablation study by removing Token Projection module and/or Contextual Sampling module on the Electricity, Traffic and Weather datasets. The best results are highlighted in \textbf{bold}.} \label{tab:ablation}
    \begin{tabular}{c|c|cc|cc|cc|cc}
    \hline
    \multicolumn{2}{c|}{Models} & \multicolumn{2}{c|}{CP-Net} & \multicolumn{2}{c|}{w/o TP} & \multicolumn{2}{c|}{w/o CS} & \multicolumn{2}{c}{w/o TP\&CS} \bigstrut\\
    \hline
    \multicolumn{2}{c|}{Metric} & MSE   & MAE   & MSE   & MAE   & MSE   & MAE   & MSE   
    & MAE \bigstrut\\
    \hline
    \multirow{4}[2]{*}{\rotatebox{90}{Electricity}} & 96    & \textbf{0.180} & \textbf{0.264} & 0.186  & 0.270  & 0.212  & 0.289  & 0.216  & 0.293  \bigstrut[t]\\
          & 192   & \textbf{0.186} & \textbf{0.269} & 0.190  & 0.275  & 0.203  & 0.285  & 0.208  & 0.290  \\
          & 336   & \textbf{0.202} & \textbf{0.286} & 0.207  & 0.291  & 0.217  & 0.299  & 0.222  & 0.304  \\
          & 720   & \textbf{0.243} & \textbf{0.319} & 0.248  & 0.324  & 0.260  & 0.332  & 0.264  & 0.337  \bigstrut[b]\\
    \hline
    \multirow{4}[2]{*}{\rotatebox{90}{Traffic}} & 96    & \textbf{0.492} & \textbf{0.325} & 0.519  & 0.344  & 0.603  & 0.399  & 0.635  & 0.412  \bigstrut[t]\\
          & 192   & \textbf{0.492} & \textbf{0.322} & 0.514  & 0.340  & 0.560  & 0.369  & 0.588  & 0.384  \\
          & 336   & \textbf{0.506} & \textbf{0.328} & 0.527  & 0.345  & 0.573  & 0.373  & 0.598  & 0.387  \\
          & 720   & \textbf{0.539} & \textbf{0.345} & 0.561  & 0.362  & 0.610  & 0.392  & 0.640  & 0.407  \bigstrut[b]\\
    \hline
    \multirow{4}[2]{*}{\rotatebox{90}{Weather}} & 96    & \textbf{0.177} & \textbf{0.217} & 0.180  & 0.219  & 0.198  & 0.240  & 0.201  & 0.245  \bigstrut[t]\\
          & 192   & \textbf{0.226} & \textbf{0.259} & 0.229  & 0.261  & 0.245  & 0.275  & 0.247  & 0.279  \\
          & 336   & \textbf{0.281} & \textbf{0.298} & 0.284  & 0.300  & 0.295  & 0.309  & 0.297  & 0.312  \\
          & 720   & \textbf{0.357} & \textbf{0.348} & 0.359  & \textbf{0.348} & 0.366  & 0.354  & 0.368  & 0.356  \bigstrut[b]\\
    \hline
    \end{tabular}%
\end{table}%

\begin{figure}[t]
    \hspace{-0.8cm}
    \includegraphics[width=1.12\textwidth]{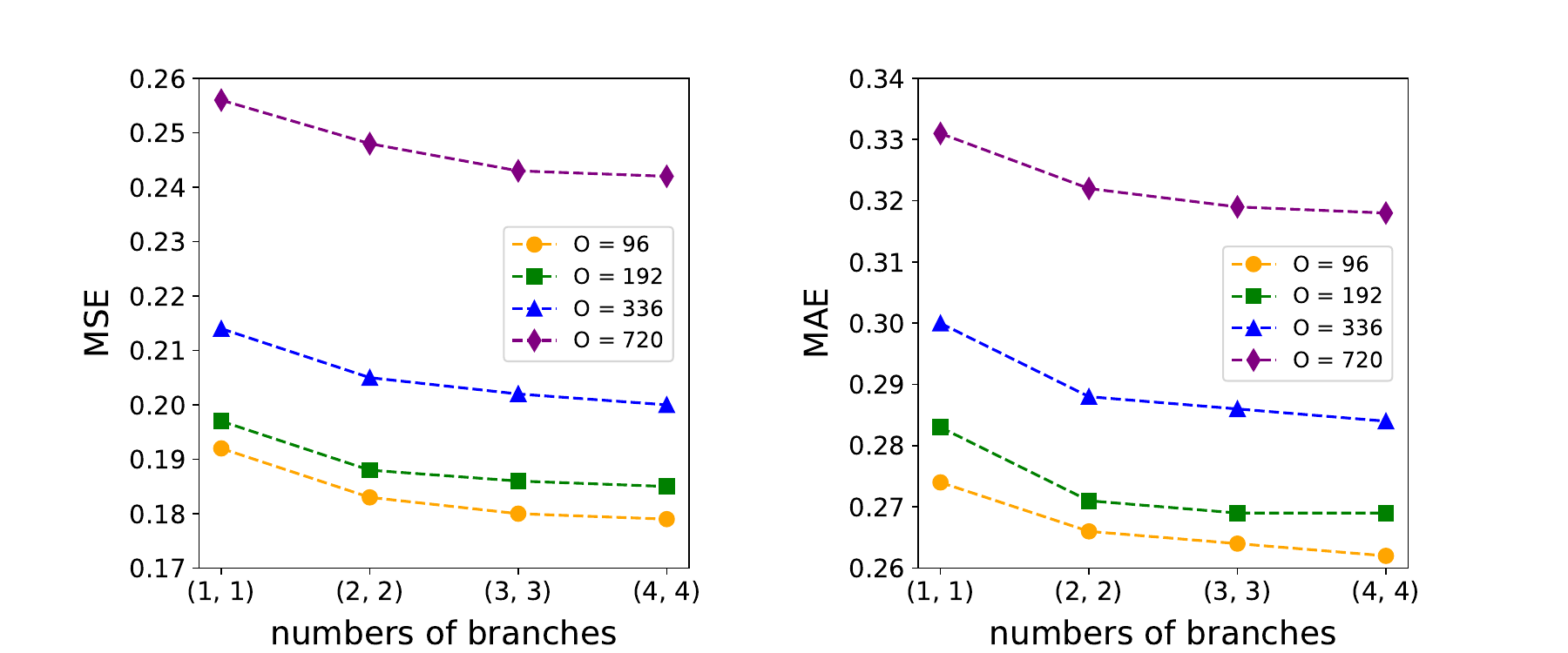}
    \caption{Impact of the number of branches on the Electricity dataset. The horizontal axis $(N_{TL}, N_{SR})$ represents the numbers of token lengths and sampling rates, respectively (for simplicity they are set to be identical).}
    \label{fig:multiscale}
\end{figure}
We carry out the ablation study on the three largest datasets, Electricity, Traffic and Weather, and the results are shown in Table \ref{tab:ablation}. Overall, the two coarsening modules together significantly improve the performance compared with the raw MLP model, with respective improvements of 10.3\%, 17.6\%, and 6.5\% in MSE on the three datasets. Using either one independently can enhance the performance of the MLP model, which demonstrate the importance of enhancing MLP model with the effective extraction of short-term patterns. Yet, it is noticeable that while incorporating the coarsening in the TP contributes to the performance, the coarsening module in the CS block exhibits a more pronounced effect. The reason may be attributed to the fact that this module not only captures clear patterns through dilated and equispaced convolutions but also form an adequate information bottleneck that eliminates noises through the down-sampling process.

Furthermore, we conduct experiments to verify the impact of the number of branches in the backbone on the performance, taking the Electricity dataset as an example. For simplicity, we set here the number of token lengths as the number of sampling rates, and we use the optimal parametric combination in each case. As illustrated in Fig. \ref{fig:multiscale}, the results show that there is a noticeable improvement as the number of branches increases in the forecasting task. However, the improvement becomes less pronounced when the number of branches reaches 4. Therefore, to balance operational efficiency and performance, we ultimately choose 3 as the number of branches.

\subsection{Model Analyses}

\subsubsection{Consistency with look-back windows.}
\begin{figure}[t]
    \hspace{-0.68cm}
    \includegraphics[width=1.07\textwidth]{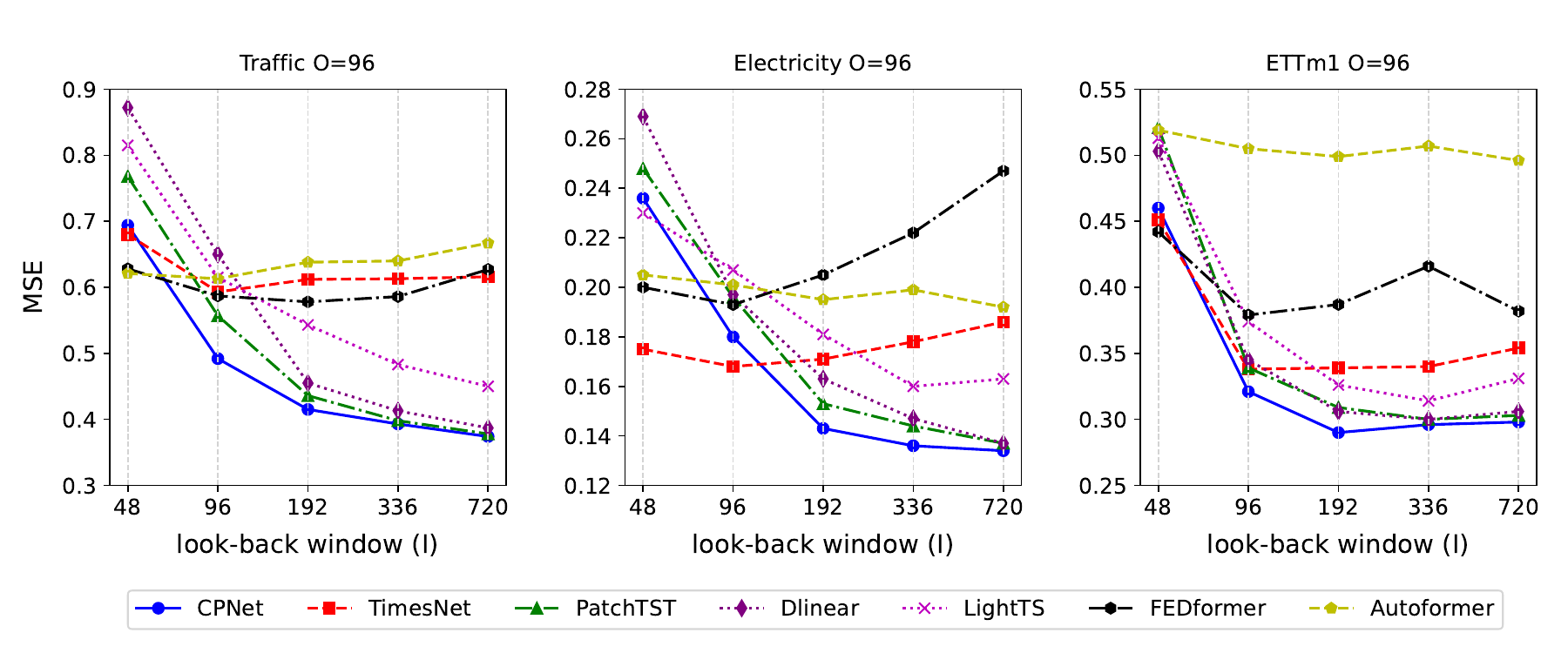}
    \caption{Forecasting performance (MSE) with varying look-back window widths $I \in \{48, 96, 192, 336, 720\}$ on the Traffic, Electricity and ETTm1 datasets. The prediction length is fixed at $O=96$. }
    \label{fig:varyingI}
\end{figure}

Previous work \cite{zeng2023transformers,nie2022time} shows that some Transformer-based models such as Autoformer and FEDformer degrade in the accuracy of prediction as the look-back window expands. With a longer look-back window, more useful information is exposed, such as long-term dependencies which cannot be captured within shorter ones. Therefore, a good time series prediction model should be able to make more accurate prediction as the look-back window expands. To further examine whether our model has this desirable ability, we train our model with different look-back windows $I \in \{48, 96, 192, 336, 720\}$ and compare its performance with other state-of-the-art models. The results are shown in Fig. \ref{fig:varyingI}. Our model performs excellently across varying look-back windows, with its performance consistently improving as the look-back window expands. Moreover, we can observe that TimesNet fails to achieve better performance when look-back window expands, which demonstrate the idea 2D convolution may still lack the essence of modeling long-term dependencies.

\subsubsection{Training and Inference Efficiency.}
\begin{figure}[t]
    \hspace{-0.45cm}
    \includegraphics[width=1.02\textwidth]{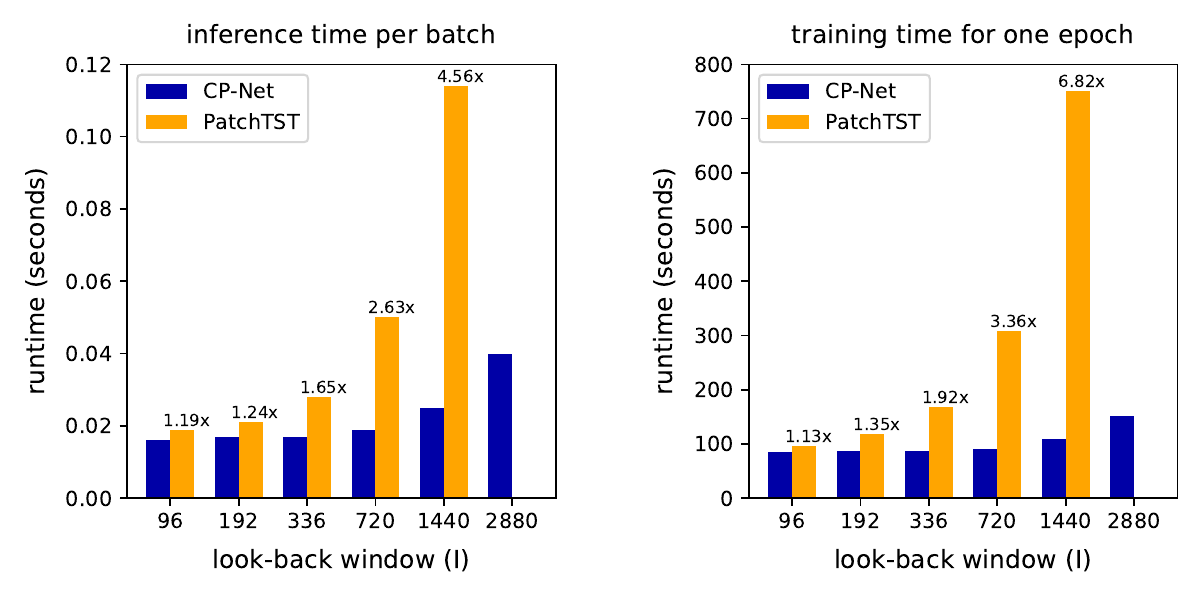}
    \caption{Comparison of training and inference time against PatchTST based on the attention mechanism as one of the state-of-art models on the Electricity dataset. Note that PatchTST encountered GPU memory exhaustion for the look-back window width $I \geq 2880$.}
    \label{fig:effiency}
\end{figure}

Beyond excelling performance compared with baseline models, we further demonstrate the high efficiency of the proposed model attributed to its MLP-based architecture. In this regard, we compare CP-Net with the SOTA model PatchTST and focus on the training and inference time. For fair comparisons, we conduct experiments using the data loader in Autoformer \cite{wu2021autoformer}. We utilize a batch size of 8 for the Electricity dataset with 321 individual time series, resulting in batches of size 8 × 321 × I (with I being the varying width of the look-back window). We report the inference time per batch and the training time for one epoch for CP-Net and PatchTST respectively as the look-back window (I) varies from 96 to 2880, as shown in Fig. \ref{fig:effiency}. Our model is significantly superior to PatchTST regarding both training and inference speeds with less fluctuation. Besides, PatchTST runs out of memory when $I \geq 2880$. It can thus be concluded that with the architectural simplicity, our model achieves a superior or comparable accuracy with improved computational and memory efficiency compared to models based on the attention mechanism. All experiments for this runtime comparison were conducted using a single NVIDIA RTX3090Ti GPU on the same machine.

\section{Conclusion}

We present the CP-Net, a novel time series forecasting model featured by an effective boosting strategy for MLPs. It employs a coarsening scheme that addresses two critical problems of the point-wise projection of the MLP layer, known as the deficient contextual dependencies and inadequate information bottleneck. At the core of the CP-Net lies a multi-scale coarsening strategy composed of Token Projection Blocks and Contextual Sampling Blocks that integrate functions prior and posterior to the MLP. By forming diverse information granules, the CP-Net exhibits two key advantages as being able to grasp essential patterns in the time series and meanwhile filtering out unwanted volatile information, both falling short with a sole MLP layer. In comparison with typical convolution-based models lacking the long-term modeling capacity and Transformer-based models with a higher computational complexity, our model is able to comprehend both local and global temporal correlations while maintains a linear computational complexity. With its architectural simplicity, our extensive experimental results show that CP-Net outperforms or is on par with the SOTA models across nearly all adopted empirical datasets, which demonstrates the effectiveness of the proposed coarsening strategy in alleviating the intrinsic problems of MLPs and enhancing their modeling capability.

%
%
%
\bibliographystyle{splncs04}  %
\bibliography{mybibliography}
%




\end{document}